\documentclass{article}

\usepackage[utf8]{inputenc} 
\usepackage[T1]{fontenc}    
\usepackage{hyperref}       
\usepackage{url}            
\usepackage{booktabs}       
\usepackage{amsfonts}       
\usepackage{nicefrac}       
\usepackage{microtype}      
\usepackage{graphicx}
\usepackage{multirow}
\usepackage{mathrsfs}
\usepackage[table,xcdraw]{xcolor}
\usepackage{colortbl}
\usepackage{pifont}
\usepackage{arxiv}

\title{A Tale of Two Experts: Cooperative Learning \\ for Source‑Free Unsupervised Domain Adaptation}

\author{
  Jiaping Yu\\
  School of Electronic Engineering\\
  Xidian University\\
  Xi'an 710071, China \\
  \texttt{jpyu@stu.xidian.edu.cn} \\
  \And
  Muli Yang \\
  Institute for Infocomm Research (I$^2$R)\\
  A*STAR \\
  Singapore \\
  \texttt{yangml@i2r.a-star.edu.sg}\\
  \And
  Jiapeng Ji \\
  School of Electronic Engineering\\
  Xidian University \\
  Xi'an 710071, China \\
  \texttt{jpji1@stu.xidian.edu.cn}\\
  \And
  Jiexi Yan \\
  School of Computer Science\\
  Xidian University\\
  Xi'an 710071, China \\
  \texttt{yanjiexi@xidian.edu.cn}\\
  \And
  Cheng Deng\\
  School of Electronic Engineering\\
  Xidian University\\
  Xi'an 710071, China \\
  \texttt{chdeng.xd@gmail.com} \\
}

\begin{document}

\maketitle

\begin{abstract}
Source‐Free Unsupervised Domain Adaptation (SFUDA) addresses the realistic challenge of adapting a source‐trained model to a target domain without access to source data, driven by privacy and cost concerns. Existing SFUDA methods either exploit only the source model's predictions or fine‐tune large multimodal models, yet both neglect complementary insights and the latent structure of target data. In this paper, we propose the Experts Cooperative Learning (EXCL). EXCL contains the Dual Experts framework and Retrieval‑Augmentation‑Interaction optimization pipeline.
The Dual Experts framework places a frozen source‐domain model (augmented with Conv‑Adapter) and a pretrained vision–language model (with trainable text prompt) on equal footing to mine consensus knowledge from unlabeled target samples. To effectively train these plug‑in modules under purely unsupervised conditions, we introduce Retrieval-Augmented-Interaction(RAIN), a three‑stage pipeline that (i) collaboratively retrieves pseudo‑source and complex target samples, (ii) separately fine‑tunes each expert on its respective sample set, and (iii) enforces learning object consistency via a shared learning result. Extensive experiments on four benchmark datasets demonstrate that our approach matches state‐of‐the‐art performance.
\end{abstract}

\begin{figure}[h]
  \centering
   \includegraphics[width=0.6\linewidth]{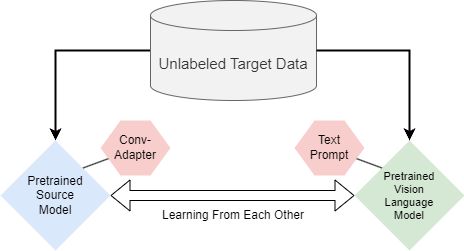}
   \caption{Illustration of Dual Expert Frameworks. The source domain data is unavailable throughout the training steps, except for the source model and unlabeled target domain data. Two expert models must learn from each other to ensure the consistency of the learning target. }
   \label{dual_experts}
\end{figure}

\section{Introduction}
\label{sec:intro}
Unsupervised Domain Adaptation (UDA) aims to address the scenario where the data distributions of the source and target domains are inconsistent and the target domain samples lack labels, by jointly training a model with labeled source domain samples and unlabeled target domain samples to achieve good generalization performance on the target domain. 
However, due to concerns over privacy, security, and cost, there is a growing call to restrict models' access to source domain training data. 
Against this backdrop, a more realistic and challenging task has emerged: Source-Free Unsupervised Domain Adaptation (SFUDA). In SFUDA, the source domain data becomes inaccessible, and the model training relies on the unlabeled samples from the target domain and the source model pre-trained on the source domain data to adapt to the target domain data.

The current methods have been divided into two major categories to address the SFUDA task. 
The first category of methods focuses on exploring how to utilize the source domain model and target domain data to enhance the generalization of the source domain model to the target domain. 
For instance, these methods~\cite{kurmi2021domain,ding2022source,tian2021vdm,tang2022semantic} primarily estimate the source domain data distribution explicitly or implicitly to construct an intermediate domain for knowledge transfer. 
These works~\cite{xia2021adaptive,huang2021model,yang2021exploiting} enhance the domain invariance of the target domain model through adversarial or contrastive feature representation. By introducing new loss functions or regularization methods, these works~\cite{kundu2022balancing,lao2021hypothesis,wang2022exploring} optimize model training on the target domain.
The second category of methods tends to fine-tune the target domain model using pre-trained multimodal models. For example,~\cite{liang2023open,tang2024source2,tang2024proxy} leverage the strong generalization ability of large multimodal models to help the source domain model adapt to the target domain dataset. 
However, these approaches typically rely solely on the source domain model for the first category of methods. 
Due to domain differences, the classification boundaries of the source domain model cannot adapt well to the feature distribution of the target domain, leading to performance degradation. 
For the second category of methods, although pre-trained models provide powerful initial representations, fine-tuning on the target domain may not fully capture the unique characteristics of the target domain, thereby limiting the model's adaptability. 
At the same time, these methods fail to adequately consider the quality, representativeness, and uncertainty of target domain data, which can lead to biases in the model adaptation process. 
Moreover, an open question is how to fully exploit the target domain data's intrinsic structure and latent information without source domain data.

In this case, we proposed a novel method called Experts Cooperative Learning, which contains the Dual Experts Framework and Retrieval-Augmented-Interaction (RAIN) optimization pipeline. 
In the Dual Experts Framework, the source domain model and the Visual Language Model (VLM) leverage their respective plug-in modules (Conv-Adapter and trainable text prompt) to mine consensual information from the target domain data, thereby achieving adaptation to the target domain dataset. 
However, unsupervised target domain data are challenging to use for targeted optimization of these plug-in modules. 
To overcome this challenge, we have designed an optimization pipeline for this framework: RAIN.
Specifically, we utilize the source domain model and the VLM to retrieve pseudo-source data information from the target domain data collaboratively. 
Subsequently, the pseudo-source and the remaining complex data are employed for targeted enhancement training of the plug-in modules. Finally, through the interaction of learning outcomes, the enhanced dual experts ensure the consistency of learning objectives, thereby improving both models' adaptation to the target domain data.

We summarize the main contributions of this paper as follows:

1. We introduce a Dual Experts framework employing a frozen source‑domain model (augmented with a Conv‑Adapter) and a pretrained VLM (with a learnable text prompt) to mine complementary consensus knowledge from unlabeled target data.

2. We design a three‑stage RAIN optimization pipeline to (a) retrieve pseudo‑source and complex target samples, (b) fine‑tune each expert’s plug‑in on these sets, and (c) enforce interactive consistency, leading to synchronized objectives and improved target‑domain adaptation.

3. We conducted extensive experiments. The results demonstrate that our method performs well with state-of-the-art (SOTA) methods in all four datasets.

\section{Related work}
\label{sec:related}
\subsection{Source-Free Unsupervised Domain Adaptation}
Existing SFUDA (Source-free Unsupervised Domain Adaptation) approaches can be categorized into two distinct types.

The first type focuses on extracting cross-domain factors from the source domain and transferring these factors through successive model adaptations to align feature distributions across the two domains. For instance, ~\cite{tang2019adaptive} establishes a mapping relationship between a sample and its corresponding individual classifier (Support Vector Machine, SVM) in the source domain to ensure effective classification in the target domain. Some approaches utilize pre-trained source models to generate auxiliary factors, including prototypes~\cite{tanwisuth2021prototype}, source distribution estimation~\cite{ding2022source}, or hard samples~\cite{li2021divergence} to facilitate feature alignment.

The second type incorporates auxiliary information derived from the unlabeled target domain. In addition to commonly used pseudo-labels~\cite{chen2022self,liang2020we}, techniques leverage geometric information~\cite{tang2021nearest,tang2022semantic}. Our method addresses this limitation by leveraging the knowledge encoded in pre-trained multimodal models and a source model for cooperative learning.

\begin{figure*}[t]
  \centering
   \includegraphics[width=0.9\linewidth]{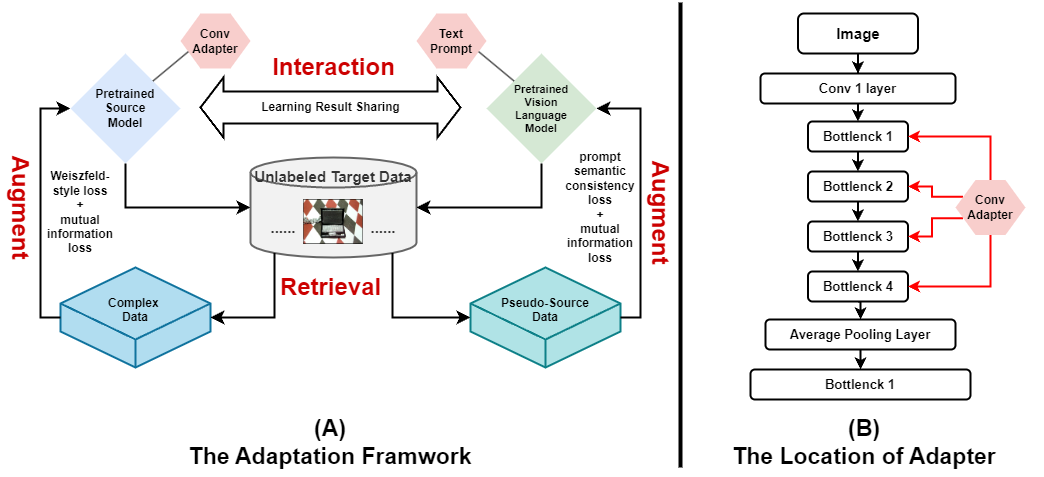}
   \caption{An overview of EXCL. Part A contains the Dual Experts Framework and the RAIN optimization Pipeline. The retrieval stage will update the pseudo-source and complex data throughout training to help the Weiszfeld Style Loss and the Promote Semantic Consistency Loss update. In the interaction stage, both experts will exchange their softmax output of the whole dataset to calculate the mutual information loss, ensuring the consistency of the learning target. Part B shows the location of the Conv-adapter plugged into the source model.}
   \label{framework}
\end{figure*}

\subsection{Pretrained Vision Language Model}
A prompt refers to a specific input pattern or textual cue given to a pre-trained model to elicit a desired response or behavior. It acts as a tool to enhance the model's performance by directing it to generate more accurate and relevant outputs based on the provided input. Depending on the model's requirements, a prompt can be formed from text or images.

Several methods focus on training a single-modal prompt for various tasks~\cite{wang2022learning, jia2022visual, lester2021power, li2021prefix, zhou2022learning, zhou2022conditional, bahng2022visual, huang2022unsupervised}. In contrast, multiple-modal prompt-tuning methods optimize prompts for each modality during model training, either in a dependent or independent manner~\cite{wang2022learning, wang2022s, liu2024compositional}. While these multiple-modal prompt-tuning methods have shown promise in enhancing model performance, they typically require supervised data, which may not be suitable for exemplar-free unsupervised domain incremental learning tasks.

Our approach optimizes a trainable text prompt using a source model equipped with an adapter and unlabeled target domain data. This allows us to perform better without relying on supervised data, effectively addressing the main challenge of source-free unsupervised domain adaptation tasks.

\section{Problem Formulation}
In the SFUDA task, the source model is trained on a labeled source domain that will be unavailable in the future and adapted to the unlabeled target domain. 
For the target domain adaptation part, the source needs to be trained on the unlabeled target domain, the model can only use the data from the current domain and has no access to the other domain data.
Formally, the labeled source domain and unlabeled target domain are both characterized by the same set of $\mathcal{C}$, the source domain data and its' corresponding labels are represented as $\mathcal{X}_s$ and $\mathcal{Y}_s$, the target domain data is defined as $\mathcal{X}_t$.
We aim to train the source domain model and the pre-trained vision language model to adapt to the target domain.
This involves utilizing several conditions: (1) a pre-trained source model $P_{s} : \mathcal{X}_s \rightarrow \mathcal{Y}_s$, (2) unlabeled target data, (3) a Pretrained Visual-Language Model $P_{v}$, and (4) the category label text set $\tau$.

\section{Methodology}
\subsection{Overview} 
As shown in Figure~\ref{framework}, the proposed method EXCL contains a two-part approach: (1) the Duel Experts framework and (2) the RAIN optimization pipeline.
In the Duel Experts framework, we elevate the role of the source-domain model, placing it on equal footing with the pretrained VLM. By introducing and optimizing a Conv-adapter and trainable text prompts, we aim to fully exploit the pretrained capacities of both expert models.
To effectively train the Duel Experts framework under limited annotations in both the source and target domains, we draw inspiration from the Retrieval-Augmented Generation (RAG) paradigm and design the RAIN pipeline. RAIN leverages the dual-expert structure to retrieve and filter target-domain samples, enhances each expert model based on the selected data, and ultimately achieves synchronized optimization through interactive training.

\subsection{Dual Experts Framework}
Within the Duel Experts framework, we elevate the status of the source domain model to match that of the VLM. 

The source domain model, trained on fully supervised data, possesses a deeper understanding of the domain-invariant and domain-specific features of the source domain, compared to the VLM, and thus qualifies as a domain expert for the source domain. 
However, the training methodology of the source domain model results in a high degree of adaptation to the source domain data, thereby restricting its generalization capability.

We introduce a Convolutional Adapter (Conv-Adapter) ~\cite{chen2024conv} to address this limitation and integrate it into the convolutional model's feature extraction module. 

Upon the introduction of the Conv-Adapter, only the parameters of the Conv-Adapter will be updated, and the parameters of the source domain model are frozen to preserve its original classification capability. 
During the training process, after the data passes through the first convolutional layer, it is simultaneously fed into the original bottleneck layer and the Conv-Adapter. 

The original bottleneck layer extracts features based on its prior training results, whereas the Conv-Adapter generates bias features targeting the features extracted by the current bottleneck layer. 
This mechanism transforms the data features of the target domain into those of the source domain, thereby mitigating the impact of the target domain's characteristics on the generalization ability of the source domain model. 
Consequently, the frozen source domain model can effectively exhibit its original performance in the target domain.

The VLM, pretrained on massive unlabeled image–text pairs, exhibits strong generalization capabilities and excels across diverse downstream tasks, thereby serving as a semantic extraction expert for the target-domain dataset.
However, the strong generalization capability of the VLM also results in suboptimal performance on specific datasets compared to specialized models. 
We introduce a trainable text prompt to address this issue~\cite{zhou2022learning}.
This text prompt can automatically learn the optimal context representation through gradient optimization, thereby replacing the original text labels and helping the VLM effectively adapt to the target domain data. 
During training, the parameters of the VLM's visual encoder and text encoder are frozen, with only the text prompt being optimized. 
This trainable text prompt enables the VLM to focus more on the semantic information in the images, reducing the impact of domain-specific features on its performance.

\subsection{RAIN optimization pipline}

The RAIN training data retrieval method comprises three key steps: retrieval, augmentation, and interaction. 

The retrieval step, which serves as the starting point of the entire process, hinges on fully leveraging the prior knowledge of the source domain model and the VLM.
The source domain model, which has been highly fitted to the source domain data through training, can accurately identify the characteristics of the source domain data. In contrast, the VLM, trained on large-scale unlabeled visual-text data pairs, demonstrates strong generalization capability in the target domain. 
By combining the source domain model's source domain data recognition capability and the VLM's zero-shot recognition capability, we can filter out a batch of data samples that can be recognized by both the source domain model and the VLM, which we refer to as pseudo-source samples.
\begin{equation}
\label{retrieval}
    P_s(x_t^i)=P_v(x_t^i) \Rightarrow x_p^i = x_t^i, x_p \in \mathcal{X}_p, x_t \in \mathcal{X}_t,
\end{equation}
Where $x_p$ is the pseudo-source samples, $x_t$ is the unlabeled target data, and $\mathcal{X}_p$ is the pseudo-source samples set.

Pseudo-source samples possess three characteristics: First, since the source domain model can recognize them, they exhibit good source domain style features. Second, as they are recognized by both the source domain model and the VLM, their semantic confidence is reasonably high. Finally, as part of the target domain dataset, they also possess target domain style features. These pseudo-source samples will be used to initialize the Conv-Adapter of the source domain model and the trainable text prompt of the VLM with the cross-entropy loss by using the common prediction result of two experts, thereby providing a good starting point for the two models to adapt to the target domain data in subsequent steps.
\begin{equation}
\label{ce_loss}
    \mathcal{L}_{ce} = -\sum^{N}_{n=1}(P_{s}(x_p^{n}), log(P_v(x_p^n))), 
\end{equation}
This cross-entropy loss will also be used in the Conv-adapter optimization in the augmentation state.

After filtering out the pseudo-source samples, the remaining target domain data are called complex samples. These complex samples, characterized by strong target domain style features, prevent the source domain model and the VLM from reaching a consensus. However, these data samples with strong target domain style features will play a crucial role in the subsequent augmentation step, supporting further model optimization.
The complex sample can be represented as:
\begin{equation}
\label{complex_data}
   \mathcal{X}_c = \mathcal{X}_t - \mathcal{X}_p,
\end{equation}
where $\mathcal{X}_c$ is the complex sample set, $x_c \in \mathcal{X}_c$ is the complex samples.

The second step is augmentation. In this step, we optimize the Conv-Adapter and the trainable text prompt using the two types of samples retrieved earlier: complex samples and pseudo-source samples.

First, we optimize the Conv-Adapter. 
We regard the target domain data as source domain data with the target domain-specific features, which causes the domain bias.
After removing the domain bias through domain-specific features alignment between the source and target domains, the distribution of the target domain data in the feature space should be consistent with that of the source domain data.
To this end, the primary role of the Conv-Adapter is to minimize the domain-specific feature difference between the target and source domains.
 This minimization aims to maximize the training effectiveness of the source domain model on the supervised dataset. 
To assist the adapter in accurately judging and learning the style differences between the target and source domains, we designed the Weiszfeld style loss.

To calculate the Weiszfeld-style loss, find each class's centers. In traditional methods, the center of a class is typically determined by calculating the mean of the feature cluster. While this approach is computationally efficient, it is sensitive to outliers. In contrast, the Weiszfeld algorithm iteratively computes the center that minimizes the sum of the Euclidean distances to all feature points in the cluster. As a result, this algorithm is less sensitive to noise and more robust.
We first use the feature extraction layer of the source domain model to project the pseudo-source samples into the feature space by category. Subsequently, based on the Weiszfeld center computation method, we obtain multiple robust and representative feature centers for each category.
The centers of representative features are calculated like this:
\begin{equation}
    y_{w+1} = (\sum^{P}_{p=1}\frac{P_s(x_{p})}{\sqrt(P_s(x_{p})^{2}-y^{2}_{w})}) / (\sum^{P}_{p=1}\frac{1}{\sqrt(P_s(x_{p})^{2}-y^{2}_{w})})
\end{equation}
$x_p$ represents the pseudo-source data,$y_{w}$ represents the current Weiszfeld center, which is initialized by the mean of $P_s(x_{p})$,$y_{w+1}$ represents the next Weiszfeld center,
These centers can calculate the style difference with the complex samples. The loss is shown as follows:
\begin{equation}
\label{weisz_loss}
	\mathcal{L}_{\mathrm{weisz}} = (1 - \frac{y_{\mathrm{w}} \cdot x_{p}}{\left\|y_{\mathrm{w}}\right\| \times \left\|x_{p}\right\|} ) + \mathcal{L}_{ce},
\end{equation}
The domain-specific feature differences can be easily captured between pseudo source domain samples and complex samples under consistent classification results.
These differences can effectively assist the adapter in learning the domain-specific feature difference from the source domain to the target domain, combined with the cross-entropy loss~\ref{ce_loss} to limit the semantic change while learning the domain-specific feature, which maximizes the performance of the source domain model on the target domain dataset..

Secondly, we optimize the Trainable Text Prompts. 
Since the VLM is not optimized for specific domain data, we introduce Trainable Text Prompts to guide the model in focusing on high-level semantic features of images rather than cross-domain style differences, thereby alleviating the degradation of classification performance caused by distribution differences between domains. Meanwhile, to ensure that the Trainable Text Prompts do not alter the core semantics of the original label text, we design a Prompt Semantic Consistency Loss to constrain the consistency of model prediction results before and after adding the prompts. 
The prompt semantic consistency loss is shown as:
\begin{equation}
    \label{psc loss}
	\mathcal{L}_{\mathrm{psc}} = \mathrm{KL}(P_{v}(x_{c},\tau) || P_{v}(x_{c}, \psi+\tau)),
\end{equation}
where the $P_{v}$ denotes the VLM model, $x_{c}$ represents the pseudo-source samples, $\psi$ represents the Trainable Text Prompts, and $\gamma$ represents the category text labels. 
This loss constrains the impact of the Trainable Text Prompts on the core semantics of the original label text by calculating the consistency of the VLM's classification probability distribution before and after adding the Trainable Text Prompts.

The interaction phase is based on the Duel Experts Framework. To enhance the prediction consistency between the two expert models, we introduce the mutual information loss \cite{ji2019invariant} between the softmax outputs of the two expert models. The objective of the mutual information loss is to maximize the mutual information between the output distributions of the two models, thereby driving the two models to evolve towards consistent predictions under unsupervised conditions.
Specifically, let the outputs of the two models be $P_s(x_t) = O_{s}$ and $P_v(x_t) = O_{v}$, and the joint distribution $O_{i,j}$ constructed based on the same batch, where $i$ and $j$ are the class indices of the current batch for the source domain model and the VLM, respectively. The joint distribution is defined as:
\begin{equation}
\label{margin_dist}
O_{i,j} = \frac{1}{N}\sum^{N}_{n=1}O^{(n)}_{s}(i) \cdot O^{(n)}_{v}(j),
\end{equation}
Where the n represents the batch size $n \in N$,
The marginal distribution of the source domain model is: $O^{s}_{i} = \sum_{j}O_{i,j}$ , the marginal distribution of the VLM is: $O^{v}_{j} = \sum_{i}O_{i,j}$.
Then the mutual information loss is:
\begin{equation}
\label{mi_loss}
 \mathcal{L}_{\mathrm{mi}}
= - \sum_{i,j} O_{i,j}\log\biggl(\frac{O_{i,j}}{O^{s}_{i} \cdot O^{v}_{j}}\biggr)).
\end{equation}
To maintain decoupled training between the two expert models, we let each model independently calculate the loss and back-propagate to update itself, using the softmax output of the other model as the alignment target. Since the mutual information loss is designed based on mutual information, it also follows the symmetric definition of mutual information, that is, $I(O^{s}_{i} \; O^{v}_{j}) = I(O^{v}_{j} \; O^{s}_{i})$. 
Therefore, in the interaction phase, the two expert models share the same mutual information loss to maintain consistency in learning objectives and optimize collaboratively towards the direction of target domain adaptation.

The $\mathcal{L}_\mathrm{weiszfeld} + \mathcal{L}_\mathrm{mi}$ used in Conv-Adapter's optimization, and $\mathcal{L}_{\mathrm{psc}} + \mathcal{L}_{\mathrm{mi}}$ used for trainable text prompt's optimization, and for these losses, there are no hyperparameters set.
These two components are interwoven and synergize to form an organic whole. The Duel Experts framework ensures the effectiveness of the data retrieved by RAIN, while the data retrieval and filtering conducted by RAIN, in turn, enhance the performance and adaptation of the Duel Experts framework to the target domain data.
\vspace{-5pt}
\section{Experiments}
\label{sec:intro}
\begin{table*}[t]
    \begin{center}
    \caption{Final Accuracy (\%) and Average Accuracy (\%) on \textbf{Offic-Home~\cite{venkateswara2017deep}} and VisDA~\cite{peng2017visda} dataset, \textbf{SF} means source-free.}
    \scalebox{0.5}{
\begin{tabular}{c|c|ccccccccccccc|c}
\hline
                         &                      & \multicolumn{13}{c|}{Office-Home}                                                                                                                                                                                                                                                                                                                                                                   & VisDA \\ \cline{3-16} 
\multirow{-2}{*}{Method} & \multirow{-2}{*}{SF} & Ar $\rightarrow$ Cl                       & Ar $\rightarrow$ Pr                       & Ar$\rightarrow$ Rw                       & Cl $\rightarrow$ Ar                       & Cl $\rightarrow$ Pr                       & Cl $\rightarrow$ Rw                       & Pr $\rightarrow$ Ar                       & Pr $\rightarrow$ Cl                       & Pr $\rightarrow$ Rw                       & Rw $\rightarrow$ Ar                       & Rw $\rightarrow$ Cl                       & Rw $\rightarrow$ Pr                       & Avg                         & Sy $\rightarrow$ Re \\ \cline{3-16} 
Source                   & -                    & 43.7                        & 67.0                        & 73.9                        & 49.9                        & 60.1                        & 62.5                        & 51.7                        & 40.9                        & 72.6                        & 64.2                        & 46.3                        & 78.1                        & 59.2                        & 49.2  \\ \hline
DAPL-R                   & \ding{56}                    & 54.1                        & 84.3                        & 84.8                        & 74.4                        & 83.7                        & 85.0                        & 74.5                        & 54.6                        & 84.8                        & 75.2                        & 54.7                        & 83.8                        & 74.5                        & 86.9  \\
PADCLIP-R                & \ding{56}                    & 57.5                        & 84.0                        & 83.8                        & 77.8                        & 85.5                        & 84.7                        & 76.3                        & 59.2                        & 85.4                        & 78.1                        & 60.2                        & 86.7                        & 76.6                        & 88.5  \\
ADCLIP-R                 & \ding{56}                    & 55.4                        & 85.2                        & 85.6                        & 76.1                        & 85.8                        & 86.2                        & 76.7                        & 56.1                        & 85.4                        & 76.8                        & 56.1                        & 85.5                        & 75.9                        & 87.7  \\
PDA-R                    & \ding{56}                    & 55.4                        & 85.1                        & 85.8                        & 75.2                        & 85.2                        & 85.2                        & 74.2                        & 55.2                        & 85.8                        & 74.7                        & 55.8                        & 86.3                        & 75.3                        & 86.4  \\
DAMP-R                   & \ding{56}                    & 59.7                        & 88.5                        & 86.8                        & 76.6                        & 88.9                        & 87.0                        & 76.3                        & 59.6                        & 87.1                        & 77.0                        & 61.0                        & 89.9                        & 78.2                        & 88.4  \\ \hline
SHOT                     & \ding{52}                    & 56.7                        & 77.9                        & 80.6                        & 68.0                        & 78.0                        & 79.4                        & 67.9                        & 54.5                        & 82.3                        & 74.2                        & 58.6                        & 84.5                        & 71.9                        & 82.7  \\
NRC                      & \ding{52}                    & 57.7                        & 80.3                        & 82.0                        & 68.1                        & 79.8                        & 78.6                        & 65.3                        & 56.4                        & 83.0                        & 71.0                        & 58.6                        & 85.6                        & 72.2                        & 85.9  \\
GKD                      & \ding{52}                    & 56.6                        & 78.2                        & 81.8                        & 68.7                        & 78.9                        & 79.1                        & 67.6                        & 54.8                        & 82.6                        & 74.4                        & 58.5                        & 84.8                        & 72.2                        & 83.0  \\
AaD                      & \ding{52}                    & 59.3                        & 79.3                        & 82.1                        & 68.9                        & 79.8                        & 79.5                        & 67.2                        & 57.4                        & 83.1                        & 72.1                        & 58.5                        & 85.4                        & 72.7                        & 88    \\
AdaCon                   & \ding{52}                    & 47.2                        & 75.1                        & 75.5                        & 60.7                        & 73.3                        & 73.2                        & 60.2                        & 454.2                       & 76.6                        & 65.6                        & 48.3                        & 79.1                        & 65.0                        & 86.8  \\
CoWA                     & \ding{52}                    & 56.9                        & 78.4                        & 81.0                        & 69.1                        & 80.0                        & 79.9                        & 67.7                        & 57.2                        & 82.4                        & 72.8                        & 60.5                        & 84.5                        & 72.5                        & 86.9  \\
ELR                      & \ding{52}                    & 58.4                        & 78.7                        & 81.5                        & 69.2                        & 79.5                        & 79.3                        & 66.3                        & 58.0                        & 82.6                        & 73.4                        & 59.8                        & 85.1                        & 72.6                        & 85.8  \\
PLUE                     & \ding{52}                    & 49.1                        & 73.5                        & 78.2                        & 62.9                        & 73.5                        & 74.5                        & 62.2                        & 48.3                        & 78.6                        & 68.6                        & 51.8                        & 81.5                        & 66.9                        & 88.3  \\
CPD                      & \ding{52}                    & 59.1                        & 79.0                        & 821.4                       & 68.5                        & 79.7                        & 79.5                        & 67.9                        & 57.9                        & 82.8                        & 73.8                        & 61.2                        & 84.6                        & 73.0                        & 85.8  \\
TPDS                     & \ding{52}                    & 59.3                        & 80.3                        & 82.1                        & 70.6                        & 79.4                        & 80.9                        & 69.8                        & 56.8                        & 82.1                        & 74.5                        & 61.2                        & 85.3                        & 73.5                        & 87.6  \\ \hline
DIFO-R                   & \ding{52}                    & 62.6                        & 87.5                        & 87.1                        & 79.5                        & 87.9                        & 87.4                        & 78.3                        & 63.4                        & 88.1                        & 80.0                        & 63.3                        & 87.7                        & 79.4                        & 88.6  \\
DIFO-V                   & \ding{52}                    & 70.6                        & 90.6                        & 88.8                        & 82.5                        & 90.6                        & 88.8                        & 80.9                        & 70.1                        & 88.9                        & {\color[HTML]{FE0000} 83.4} & 70.5                        & 91.2                        & 83.1                        & 90.3  \\
ProDe-R                  & \ding{52}                    & 64.0                        & 90.0                        & 88.3                        & 81.1                        & 90.1                        & 88.6                        & 79.8                        & 65.4                        & 89.0                        & 80.9                        & 65.5                        & 90.2                        & 81.1                        & 88.7  \\
ProDe-V                  & \ding{52}                    & 72.7                        & 92.3                        & 90.5                        & 82.5                        & 91.5                        & 90.7                        & 82.5                        & {\color[HTML]{FE0000} 72.5} & 90.8                        & 83.0                        & {\color[HTML]{FE0000} 72.6} & 92.2                        & 84.5                        & 91.0  \\
\rowcolor[HTML]{C0C0C0} 
Our-R                    & \ding{52}                    & 67.2                        & 90.51                       & 88.45                       & 79.65                       & 91.09                       & 89.60                       & 81.59                       & 68.28                       & 89.86                       & 81.33                       & 68.5                        & 88.46                       & 82.0                        & 88.8  \\
\rowcolor[HTML]{C0C0C0} 
Our-V                    & \ding{52}                    & {\color[HTML]{FE0000} 72.3} & {\color[HTML]{FE0000} 92.1} & {\color[HTML]{FE0000} 91.3} & {\color[HTML]{FE0000} 83.9} & {\color[HTML]{FE0000} 92.5} & {\color[HTML]{FE0000} 90.9} & {\color[HTML]{FE0000} 83.4} & 72.1                        & {\color[HTML]{FE0000} 91.0} & 83.3                        & 72.1                        & {\color[HTML]{FE0000} 93.1} & {\color[HTML]{FE0000} 84.8} & {\color[HTML]{FE0000}91.2}  \\ \hline
\end{tabular}}
    \label{table1}
    \end{center}
\end{table*}
In this section, we validate our proposed method on four benchmark datasets and compare and analyze the results with other methods.
\subsection{Experiment Setup}
\textbf{Datasets} We have chosen three commonly used benchmark datasets to simulate the source-free unsupervised domain adaptation scenarios.These datasets are: DomainNet-126~\cite{saito2019semi}, Office-Home~\cite{venkateswara2017deep}, Office-31~\cite{saenko2010adapting} and VisDA~\cite{peng2017visda}.




\textbf{Implementation Details.} 
Unless stated otherwise, we implement our proposed method in PyTorch using a single Nvidia RTX A6000 GPU. We implement the source domain model with ResNet50~\cite{he2016deep}, and the VLM utilizes the image encoder architecture of ViT-B/32~\cite{dosovitskiy2010image} and the text encoder architecture of CLIP~\cite{radford2021learning}for our proposed method across all tasks and domains. The total class number in the experiments is 31 for Office-31~\cite{saenko2010adapting}, 65 for Office-Home~\cite{venkateswara2017deep}, 12 for VisDA~\cite{peng2017visda}, and 126 for DomainNet-126~\cite {saito2019semi}. The initiation of the trainable text prompt is "a photo of a [CLASS]". 
For the easy use of ViT~\cite{dosovitskiy2010image} and CLIP~\cite{radford2021learning}, the seeds of each training round are frozen, and the embedding dimension has been set to 768.

\textbf{Training Details.}
Our method trains our model using the SGD optimizer with a momentum of 0.9. The initial learning rate of the Conv-Adapter of the source domain model and the trainable text prompt of the VLM have been set to 0.1 and 0.01, respectively. 
The batch size for training has been established at 64, and the number of training epochs differs depending on the dataset. 
All images from the three datasets were resized to 224x224 and augmented through horizontal flipping and random cropping.
\begin{table}[]
\centering
 \caption{Final Accuracy (\%) and Average Accuracy (\%) of Closed-set SFUDA on \textbf{DomainNet-126~\cite{saito2019semi}} dataset, \textbf{SF} means source-free}
    \scalebox{0.65}{
\begin{tabular}{c|c|ccccccccccccc}
\hline
                         &                      & \multicolumn{13}{c}{DomainNet-126}                                                                                                                                                                                                                                                                                                                                                                  \\
\multirow{-2}{*}{Method} & \multirow{-2}{*}{SF} & C $\rightarrow$ P                         & C $\rightarrow$ R                         & C $\rightarrow$ R                         & P $\rightarrow$ C                         & P $\rightarrow$ R                         & P $\rightarrow$ S                         & R $\rightarrow$ C                         & R $\rightarrow$ P                         & R $\rightarrow$ S                         & S $\rightarrow$ C                         & S $\rightarrow$ P                         & S $\rightarrow$ R                         & Avg                         \\ \hline
Source                   & -                    & 44.6                        & 59.8                        & 47.5                        & 53.3                        & 75.3                        & 46.2                        & 55.3                        & 62.7                        & 46.4                        & 55.1                        & 50.7                        & 59.5                        & 54.7                        \\ \hline
DAPL-R                   & \ding{56}                    & 72.4                        & 87.6                        & 65.9                        & 72.7                        & 87.6                        & 65.6                        & 73.2                        & 72.4                        & 66.2                        & 73.8                        & 72.9                        & 87.8                        & 74.8                        \\
ADCLIP-R                 & \ding{56}                    & 71.7                        & 88.1                        & 66.0                        & 73.2                        & 86.9                        & 65.2                        & 73.6                        & 73.0                        & 68.4                        & 72.3                        & 74.2                        & 89.3                        & 75.2                        \\
DAMP-R                   & \ding{56}                    & 76.7                        & 88.5                        & 71.7                        & 74.2                        & 88.7                        & 70.8                        & 74.4                        & 75.7                        & 70.5                        & 74.9                        & 76.1                        & 88.2                        & 77.5                        \\ \hline
SHOT                     & \ding{52}                    & 63.5                        & 78.2                        & 59.5                        & 67.9                        & 81.3                        & 61.7                        & 67.7                        & 67.6                        & 57.8                        & 70.2                        & 64.0                        & 78.0                        & 68.1                        \\
GKD                      & \ding{52}                    & 61.4                        & 77.4                        & 60.3                        & 69.6                        & 81.4                        & 63.2                        & 68.3                        & 68.4                        & 59.5                        & 71.5                        & 65.2                        & 77.6                        & 68.7                        \\
NRC                      & \ding{52}                    & 62.6                        & 77.1                        & 58.3                        & 62.9                        & 81.3                        & 60.7                        & 64.7                        & 69.4                        & 58.7                        & 69.4                        & 65.8                        & 78.7                        & 67.5                        \\
AdaCon                   & \ding{52}                    & 60.8                        & 74.8                        & 55.9                        & 62.2                        & 78.3                        & 58.2                        & 63.1                        & 68.1                        & 55.6                        & 67.1                        & 66.0                        & 75.4                        & 65.4                        \\
CoWA                     & \ding{52}                    & 64.6                        & 80.6                        & 60.6                        & 66.2                        & 79.8                        & 60.8                        & 69.0                        & 67.2                        & 60.0                        & 69.0                        & 65.8                        & 79.9                        & 68.6                        \\
PLUE                     & \ding{52}                    & 59.8                        & 74.0                        & 56.0                        & 61.6                        & 78.5                        & 57.9                        & 61.6                        & 65.9                        & 53.8                        & 67.5                        & 64.3                        & 76.0                        & 64.7                        \\
TPDS                     & \ding{52}                    & 62.9                        & 77.1                        & 59.8                        & 65.6                        & 79.0                        & 61.5                        & 66.4                        & 67.0                        & 58.2                        & 68.6                        & 64.3                        & 75.3                        & 67.1                        \\ \hline
DIFO-R                   & \ding{52}                    & 73.8                        & 89.0                        & 69.4                        & 74.0                        & 88.7                        & 79.1                        & 74.8                        & 74.6                        & 69.6                        & 74.7                        & 74.3                        & 88.0                        & 76.7                        \\
DIFO-V                   & \ding{52}                    & 76.6                        & 87.2                        & 74.9                        & 80.0                        & 87.4                        & 75.6                        & 80.8                        & 77,3                        & 75.5                        & 80.5                        & 76.7                        & 87.3                        & 80.0                        \\
ProDe-R                  & \ding{52}                    & 79.3                        & 91.0                        & 75.3                        & 80.0                        & 90.9                        & 75.6                        & 80.4                        & 78.9                        & 75.4                        & 80.4                        & 79.2                        & 91.0                        & 81.5                        \\
ProDe-V                  & \ding{52}                    & 83.2                        & 92.4                        & {\color[HTML]{FE0000} 79.0} & 85.0                        & 92.3                        & {\color[HTML]{FE0000} 79.3} & 85.5                        & 83.1                        & {\color[HTML]{FE0000} 79.1} & 85.5                        & {\color[HTML]{FE0000} 83.4} & 92.4                        & 85.0                        \\
\rowcolor[HTML]{C0C0C0} 
Our-R                    & \ding{52}                    & 80.9                        & 91.1                        & 74.2                        & 80.0                        & 91.4                        & 74.7                        & 82.3                        & 78.7                        & 75.0                        & 80.6                        & 78.5                        & 92.0                        & 81.6                        \\
\rowcolor[HTML]{C0C0C0} 
Our-V                    & \ding{52}                    & {\color[HTML]{FE0000} 83.6} & {\color[HTML]{FE0000} 92.6} & 78.9                        & {\color[HTML]{FE0000} 85.6} & {\color[HTML]{FE0000} 93.0} & {\color[HTML]{FE0000} 79.3} & {\color[HTML]{FE0000} 85.6} & {\color[HTML]{FE0000} 84.3} & 78.9                        & {\color[HTML]{FE0000} 84.8} & 82.9                        & {\color[HTML]{FE0000} 93.1} & {\color[HTML]{FE0000} 85.2} \\ \hline
\end{tabular}}
    \label{table2}
\end{table}

\textbf{Compared Methods.}
We consider three SFUDA settings: Close-set, Partial-set, and Open-set SFUDA, with a selection of 17 related comparisons.
We compared our proposed method with several others and categorized them into three groups: Classic SFUDA methods, VLM-guiding-based SFUDA methods, and Source model + VLM SFUDA methods.

In the classic SFUDA method category, we chose SHOT~\cite{liang2020we}, NRC~\cite{yang2021exploiting}, GKD~\cite{tang2021model}, HCL~\cite{huang2021model}, AaD~\cite{yang2022attracting}, AdaCon~\cite{chen2022contrastive}, CoWA~\cite{lee2022confidence}, ELR~\cite{yi2023source}, PLUE~\cite{litrico2023guiding}, and TPDS~\cite{tang2024source1}. In the VLM-guiding-based SFUDA methods, we chose DAPL~\cite{ge2023domain}, PADCLIP~\cite{lai2023padclip}, and ADCLIP~\cite{singha2023ad}. In the source model + VLM SFUDA methods, we chose DIFO~\cite{tang2024source2} and ProDe~\cite{tang2024proxy}.



\textbf{Evaluation Metrics.}
Two accuracy measurements have been used to evaluate the performance of our proposed method and others. 

(1). Average Accuracy: The average classification accuracy of the source-to-target task in source-free unsupervised domain adaptation.

(2). Final Accuracy: The final classification accuracy of each source-to-target task in source-free unsupervised domain adaptation.

\begin{figure*}[h]
	\centering
	\includegraphics[scale=0.4]{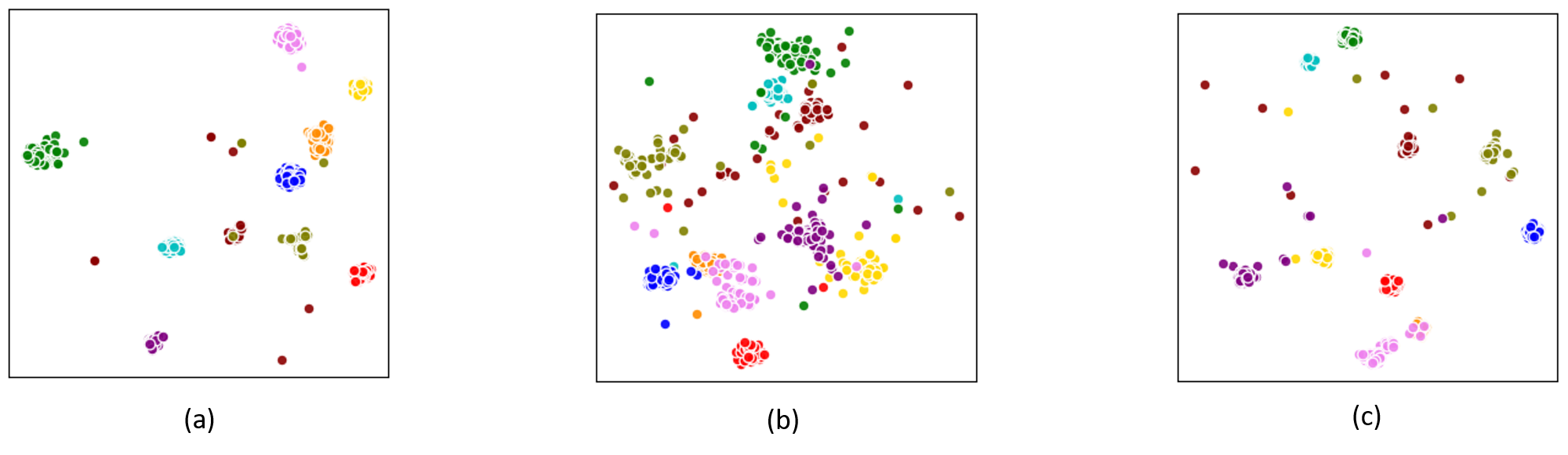}
	\caption{The t-SNE visualization shows ten classes from the Office-Home~\cite{venkateswara2017deep} dataset at the test stage, including all trained domain data. Three figures represent the source model processed source domain image feature, the source model processed target domain image feature, and the source model with the plugged adapter process target domain image feature.}
	\label{tsne}
\end{figure*}
\begin{table}[]
\centering
\caption{Final Accuracy (\%) and Average Accuracy (\%) on \textbf{Offic-31~\cite{saenko2010adapting}} dataset.}
    \scalebox{0.5}{
\begin{tabular}{ccccccccccccccccccc}
\hline
\multicolumn{19}{c}{Office-31}                                                                                                                                                                                                                                                                                                                                                                                                                                                   \\ \hline
\multicolumn{1}{c|}{Method} & \multicolumn{1}{c|}{Source} & SHOT                        & NRC                         & GKD                         & HCL                         & AaD                         & AdaCon & CoWA & ELR  & PLUE & CPD                         & \multicolumn{1}{c|}{TPDS} & DIFO-R & DIFO-R                      & ProDe-R & \cellcolor[HTML]{FFFFFF}ProDe-R & \cellcolor[HTML]{C0C0C0}Our-R & \cellcolor[HTML]{C0C0C0}Our-R                       \\ \hline
\multicolumn{1}{c|}{SF}     & \multicolumn{1}{c|}{-}      & \ding{52}                           & \ding{52}                           & \ding{52}                           & \ding{52}                           & \ding{52}                           & \ding{52}      & \ding{52}    & \ding{52}    & \ding{52}    & \ding{52}                           & \multicolumn{1}{c|}{\ding{52}}    & \ding{52}      & \ding{52}                           & \ding{52}       & \ding{52}                               & \cellcolor[HTML]{C0C0C0}\ding{52}     & \cellcolor[HTML]{C0C0C0}\ding{52}                           \\ \hline
\multicolumn{1}{c|}{A $\rightarrow$ D}    & \multicolumn{1}{c|}{79.1}   & 93.7                        & 96.0                        & 94.6                        & 94.7                        & 96.4                        & 87.7   & 94.4 & 93.8 & 89.2 & 96.6                        & \multicolumn{1}{c|}{97.1} & 93.6   & {\color[HTML]{FE0000} 97.2} & 94.4    & 96.8                            & \cellcolor[HTML]{C0C0C0}93.2  & \cellcolor[HTML]{C0C0C0}96.6                        \\
\multicolumn{1}{c|}{A $\rightarrow$ W}    & \multicolumn{1}{c|}{76.6}   & 91.1                        & 90.8                        & 91.6                        & 92.5                        & 92.1                        & 83.1   & 95.2 & 93.3 & 88.4 & 94.2                        & \multicolumn{1}{c|}{94.5} & 92.1   & 95.5                        & 92.1    & {\color[HTML]{FE0000} 96.4}     & \cellcolor[HTML]{C0C0C0}92.8  & \cellcolor[HTML]{C0C0C0}95.7                        \\
\multicolumn{1}{c|}{D $\rightarrow$ A}    & \multicolumn{1}{c|}{59.9}   & 74.2                        & 75.3                        & 75.1                        & 75.9                        & 75.0                        & 73.7   & 76.2 & 76.2 & 72.8 & 77.3                        & \multicolumn{1}{c|}{75.7} & 78.5   & 83.0                        & 79.8    & 83.1                            & \cellcolor[HTML]{C0C0C0}81.5  & \cellcolor[HTML]{C0C0C0}{\color[HTML]{FE0000} 84.1} \\
\multicolumn{1}{c|}{D $\rightarrow$ W}    & \multicolumn{1}{c|}{95.5}   & 98.2                        & 99.0                        & 98.7                        & 98.2                        & {\color[HTML]{FE0000} 99.1} & 91.3   & 98.5 & 98.0 & 97.1 & 98.2                        & \multicolumn{1}{c|}{98.7} & 95.7   & {\color[HTML]{FE0000} 97.2} & 95.6    & 97.0                            & \cellcolor[HTML]{C0C0C0}93.8  & \cellcolor[HTML]{C0C0C0}97.0                        \\
\multicolumn{1}{c|}{W $\rightarrow$ A}    & \multicolumn{1}{c|}{61.4}   & 74.6                        & 74.6                        & 75.0                        & 75.1                        & 77.7                        & 76.5   & 77.6 & 76.9 & 69.6 & 78.3                        & \multicolumn{1}{c|}{75.5} & 78.8   & 83.2                        & 79.0    & 82.5                            & \cellcolor[HTML]{C0C0C0}82.0  & \cellcolor[HTML]{C0C0C0}{\color[HTML]{FE0000} 83.4} \\
\multicolumn{1}{c|}{W $\rightarrow$ D}    & \multicolumn{1}{c|}{98.8}   & {\color[HTML]{FE0000} 100.} & {\color[HTML]{FE0000} 100.} & {\color[HTML]{FE0000} 100.} & {\color[HTML]{FE0000} 100.} & {\color[HTML]{FE0000} 100.} & 72.8   & 99.8 & 100. & 97.9 & {\color[HTML]{FE0000} 100.} & \multicolumn{1}{c|}{99.8} & 97.0   & 98.8                        & 98.6    & 99.8                            & \cellcolor[HTML]{C0C0C0}99.8  & \cellcolor[HTML]{C0C0C0}{\color[HTML]{FE0000} 100}  \\ \hline
\multicolumn{1}{c|}{Avg.}   & \multicolumn{1}{c|}{78.6}   & 88.6                        & 89.4                        & 89.2                        & 89.8                        & 89.9                        & 81.0   & 90.3 & 89.6 & 85.8 & 90.9                        & \multicolumn{1}{c|}{90.2} & 89.3   & 92.5                        & 89.9    & 92.6                            & \cellcolor[HTML]{C0C0C0}90.5  & \cellcolor[HTML]{C0C0C0}{\color[HTML]{FE0000} 92.8} \\ \hline
\end{tabular}}
    \label{table3}
\end{table}

\subsection{Main Result}
Comparison with Existing Methods
This section compares our proposed method with the representative approaches mentioned earlier. We report the Final and Average accuracy under the closed-set, partial-set, and open-set SFUDA settings. The results are presented in Table~\ref{table1}, Table~\ref{table2}, Table~\ref{table3}, and Table~\ref{table4}.
We begin by comparing the final accuracy of our method with all previous approaches. As shown in Table~\ref{table1}, Table~\ref{table2}, Table~\ref{table3}, and Table~\ref{table4}, our proposed method consistently reaches the state-of-the-art performance across all tasks and datasets.

\textbf{(1) Comparison with Classic SFUDA Methods.}

Traditional SFUDA methods rely solely on the pretrained source model and unlabeled target domain data for adaptation. Without external guidance or auxiliary information, these methods are more susceptible to noise in the target domain, undermining their ability to learn robust and transferable features. This limitation is evident in the relatively lower accuracy reported in all four tables compared to our method.

\textbf{(2) Comparison with VLM-Guided SFUDA Methods.}

VLM-guided methods leverage the generalization capabilities of large-scale vision-language models to assist target domain adaptation. However, these models are pretrained on massive web-scale image–text pairs and are not tailored for specific domain adaptation tasks. Although prompt engineering can partially steer VLMs toward downstream objectives, their performance remains constrained by the noise inherent in unlabeled target data. As a result, their adaptation effectiveness is often limited, as reflected in several tasks where our method surpasses theirs.

\textbf{(3) Comparison with Source Model + VLM Methods.}

We compare our method with approaches integrating the source model and VLM, such as DIFO and ProDe. DIFO adopts a teacher–student paradigm, where the VLM serves as the teacher to distill knowledge into a target model initialized from the source model. It also introduces pseudo-label construction based on neighborhood similarity to leverage unlabeled target data. However, the pseudo-labels generated still contain noise, which may hinder effective optimization of the target model. The performance drop observed on multiple datasets confirms this issue.

ProDe, on the other hand, introduces a controllable noise mechanism to guide the target model jointly and VLM during training, thereby mitigating the impact of noisy target samples. Nevertheless, ProDe still treats the source-initialized target model as a subordinate learner under VLM supervision. In contrast, our method adopts a cooperative learning perspective, where the source model and VLM are regarded as equal-status experts. Through mutual enhancement and consistency-driven optimization, both experts collaboratively adapt to the target domain. The competitive results of our approach—comparable to or even outperforming ProDe across all four datasets—demonstrate that the source model can serve not only as an initialization but also as a strong and autonomous expert in the adaptation process.
\vspace{-5pt}
\subsection{Ablation Study}
To evaluate the effectiveness of various components of our proposed methods, we conducted a quantitative test by selecting different loss functions to optimize the model. As shown in Table~\ref{loss_ablation}, three loss functions contribute to the training performance of auto prompts for images.

First, we verified that the mutual information loss establishes a solid foundation for the training. 
Second, by utilizing Weiszfeld Style Loss and Prompt Semantic Consistency Loss, we demonstrated the effectiveness of targeted augmented for the Conv-adapter and trainable text prompts, which enhance the source model adaptation on the target domain.
The cooperative of the source model and VLM appropriately captures intra-domain and inter-domain relationships, enhances the acquisition of domain-specific information, and improves the differences in adapters across various domains. Overall, this combination of losses achieves the best performance.
\begin{table}[]
    \begin{minipage}[t]{0.45\textwidth}
        \centering
        \caption{Average Accuracy (\%) of Partial-set \& Open-set SFUDA on Office-Home~\cite{venkateswara2017deep}}
        \scalebox{0.65}{
         \begin{tabular}{cc|cc}
\hline
Partial-set SFUDA & Avg. & Open-set SFUDA & Avg. \\ \hline
Source            & 62.8 & Source         & 46.6 \\ \hline
SHOT              & 79.3 & SHOT           & 72.8 \\
HCL               & 79.6 & HCL            & 72.6 \\
CoWA              & 83.2 & CoWA           & 73.2 \\
AaD               & 79.7 & AaD            & 71.8 \\
CRS               & 80.6 & CRS            & 73.2 \\
DIFO-V            & 84.1 & DIFO-V         & 75.9 \\
ProDe-V           & 84.2 & ProDe-V        & 82.6 \\
COL-V             & 84.0 & COL-V          & 82.9 \\ \hline
\end{tabular}}  
\label{table4}
    \end{minipage}
    \begin{minipage}[t]{0.45\textwidth}
        \centering
     \caption{Ablation studies for losses. Average Accuracy (\%) has been used on dataset \textbf{Office-Home~\cite{venkateswara2017deep}}, Office-31~\cite{saenko2010adapting}, and VisDA~\cite{peng2017visda} to measure the performance.}
    \scalebox{0.6}{
        \begin{tabular}{ccc|ccc|c}
\hline
$\mathcal{L}_{weisz}$ & $\mathcal{L}_{psc}$ & $\mathcal{L}_{mi}$ & Office-31                   & Office-Home                 & VisDA                       & Avg.                        \\ \hline
\ding{56}     & \ding{56}     & \ding{56}     & 78.6                        & 59.2                        & 49.2                        & 62.3                        \\
\ding{56}     & \ding{56}     & \ding{52}     & 81.6                        & 78.5                        & 83.9                        & 81.3                        \\
\ding{56}     & \ding{52}     & \ding{56}     & 78.3                        & 59.0                        & 49.2                        & 62.3                        \\
\ding{52}     & \ding{56}     & \ding{56}     & 78.7                        & 76.3                        & 77.6                        & 77.5                        \\
\ding{56}     & \ding{52}     & \ding{52}     & 80.4                        & 79.7                        & 86.3                        & 82.1                        \\
\ding{52}     & \ding{56}     & \ding{52}     & 85.5                        & 80.5                        & 87.1                        & 84.3                        \\
\ding{52}     & \ding{52}     & \ding{52}     & {\color[HTML]{FE0000} 92.8} & {\color[HTML]{FE0000} 84.8} & {\color[HTML]{FE0000} 91.2} & {\color[HTML]{FE0000} 89.6} \\ \hline
\end{tabular}}
        \label{loss_ablation}
    \end{minipage}
    \label{tabel4}
\end{table}

\vspace{-5pt}
\subsection{Visualizations}
We present visualizations in Fig.~\ref{tsne} of the output spaces generated by our proposed method, COL, across the source and target domains to facilitate intuitive comparison. The examples we selected for visualization are from the Office-Home~\cite{venkateswara2017deep} dataset.

The image part (a) shows the extracted features that the source model processed in the source domain images, which conform to a well-clustered distribution typical of supervised training.
The image part (b) shows the extracted features the source model processed in the target domain images. 
Due to the high degree of fitting of the source domain model to the source domain data, the features extracted from the target domain images by the source domain model exhibit poor class separability.
The image part (c) shows the extracted features that the source model, with the target domain conv-adapter, processed in the target domain images.
The Conv-Adapter effectively aligns the target domain features with the source domain distribution, making it recognizable by the source domain model, thereby ensuring that the source domain model achieves performance on the target domain data comparable to that obtained through supervised training.

\vspace{-5pt}
\section{Conclusion}
This paper introduces Experts Cooperative Learning(EXCL), a novel approach for Source-Free Unsupervised Domain Adaptation (SFUDA), which integrates the Dual Experts framework and the Retrieval-Augmentation-Interaction (RAIN) optimization pipeline.
The Dual Experts framework combines a frozen source-domain model augmented with a Conv-Adapter and a pre-trained vision–language model equipped with a learnable text prompt. The proposed framework leverages the complementary strengths of these models to mine consensus knowledge from unlabeled target samples. 
The RAIN pipeline effectively trains the plug-in modules by retrieving pseudo-source and complex target samples, fine-tuning each expert separately, and enforcing learning objective consistency. Extensive experiments on four benchmark datasets demonstrate. Numerical analysis and performance measurement studies demonstrate the superiority and effectiveness of EXCL in handling SFUDA.

{
    \small
    \bibliographystyle{plain}
    \bibliography{neurips_2025}
}

\end{document}